\documentclass{article}

\usepackage{arxiv}

\usepackage[utf8]{inputenc} 
\usepackage[T1]{fontenc}    
\usepackage{hyperref}       
\usepackage{url}            
\usepackage{booktabs}       
\usepackage{amsfonts}       
\usepackage{nicefrac}       
\usepackage{microtype}      
\usepackage{lipsum}
\usepackage{graphicx}
\usepackage{epsfig}
\usepackage{subcaption}
\usepackage{xcolor}
\usepackage{pifont}
\usepackage{amsmath,amssymb} 
\usepackage{multirow}
\usepackage{upgreek}
\usepackage{arydshln}
\usepackage{tikz}
\usepackage{comment}
\usepackage{color}
\DeclareMathAlphabet{\pazocal}{OMS}{zplm}{m}{n}

\title{Arbitrary Style Guidance for Enhanced Diffusion-Based Text-to-Image Generation}

\author{
  Zhihong Pan, Xin Zhou, Hao Tian \\
  Baidu Research (USA)\\
   \\
}

\begin{document}
\newcommand{\cmark}{\ding{51}}%
\newcommand{\xmark}{\ding{55}}%
\newcommand{\red}[1]{\textcolor{red}{#1}}
\newcommand{\blue}[1]{\textcolor{blue}{#1}}
\newcommand{\green}[1]{\textcolor{green}{#1}}
\newcommand{\teal}[1]{\textcolor{teal}{#1}}
\newcommand{\orange}[1]{\textcolor{orange}{#1}}
\newcommand{\wt}[1]{\textcolor{white}{#1}}
\newcommand{\etal}{\textit{et al.}}

\maketitle

\begin{abstract}
Diffusion-based text-to-image generation models like GLIDE and DALLE-2 have gained wide success recently for their superior performance in turning complex text inputs into images of high quality and wide diversity.  In particular, they are proven to be very powerful in creating graphic arts of various formats and styles.  Although current models supported specifying style formats like oil painting or pencil drawing, fine-grained style features like color distributions and brush strokes are hard to specify as they are randomly picked from a conditional distribution based on the given text input.  Here we propose a novel style guidance method to support generating images using arbitrary style guided by a reference image.  The generation method does not require a separate style transfer model to generate desired styles while maintaining image quality in generated content as controlled by the text input.  Additionally, the guidance method can be applied without a style reference, denoted as self style guidance, to generate images of more diverse styles.  Comprehensive experiments prove that the proposed method remains robust and effective in a wide range of conditions, including diverse graphic art forms, image content types and diffusion models.
\end{abstract}

\vspace{-5pt}
\section{Introduction}
\label{sec:intro}

Various types of deep generative models have been developed in recent years for various applications in content generation and artistic creations.
Among them, models based on Generative Adversarial Network (GAN)~\cite{goodfellow_nips_2014} have been the most successful ones for their ability
to create high quality contents with fast sampling speed~\cite{donahue_arxiv_2018, tulyakov_cvpr_2018, karras_cvpr_2019, karras_nips_2021}.
However, they have their own limitations in diversity and training stability.  Recently, denoising diffusion
models~\cite{sohl_icml_2015, ho_nips_2020, song_iclr_2021} have gained popularity increasingly for their advantages in generating images
with high qualities in both fidelity and diversity.  In addition to image generation, diffusion models have also shown successful applications in other
data modalities like 3D point clouds~\cite{luo_cvpr_2021}, audio~\cite{kong_iclr_2021} and video~\cite{ho_iclrw_2022}.
For the most popular image generation task, it has been utilized in a broad range of applications, including image-to-image translation~\cite{saharia_nipsw_2021, avrahami_cvpr_2022},
image super-resolution~\cite{saharia_arxiv_2021, wang_arxiv_2021}, image editing~\cite{meng_arxiv_2021} and image inpainting~\cite{lugmayr_cvpr_2022, romero_cvpr_2022}.
It has also empowered the breakthrough developments in diffusion-based text-to-image
models~\cite{rombach_cvpr_2022, nichol_arxiv_2022, ramesh_arxiv_2022, saharia_arxiv_2022} which are able to create realistic images according
to given text descriptions, even long and complex ones.

 \begin{figure}[t]
 \begin{center}
     \includegraphics[width=0.8\linewidth]{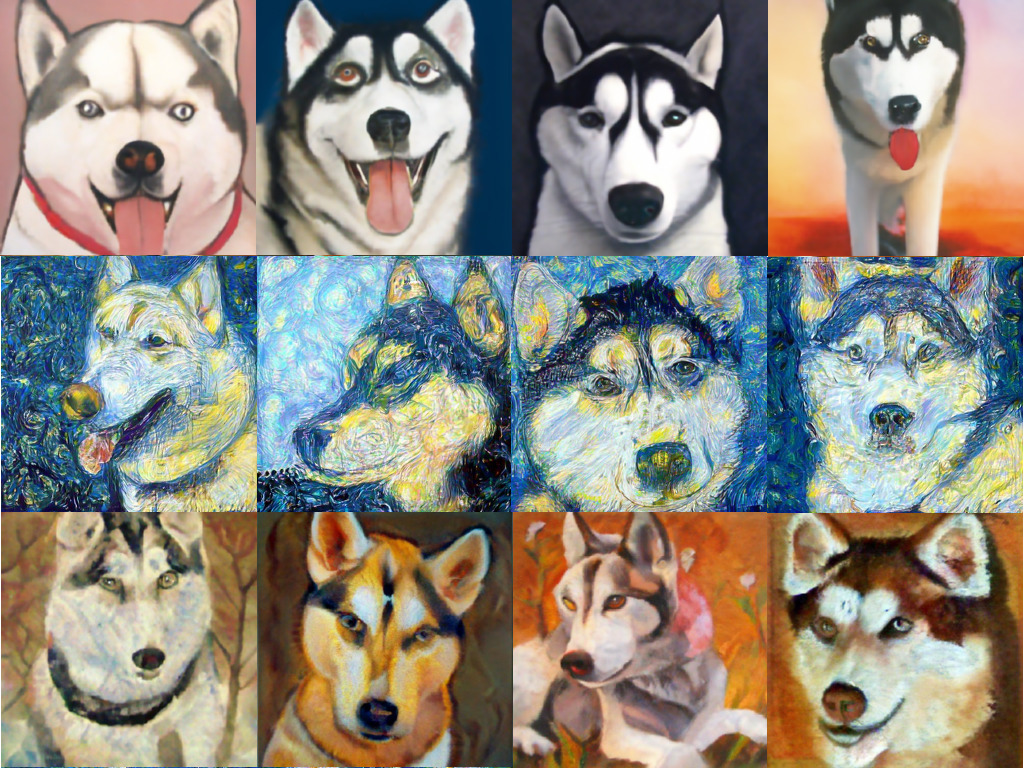}
 \end{center}
 \vspace{-6pt}
 \caption[Caption for LOF]{Comparison of three sampling methods (four samples each) using a prompt of "\textit{an oil painting of a husky}": 1) Unguided generation produces similar styles; 2) Style guided generation using van Gogh's The Starry Night; 3) Self style guidance shows samples of diverse styles\protect\footnotemark.}
 \vspace{-6pt}
 \label{fig:husky}
 \end{figure}
 \footnotetext{All examples are sampled from OpenAI's filterd GLIDE model: \url{https://github.com/openai/glide-text2im}}

A main type of content generated by these models is graphic artworks with contents match
with the corresponding text inputs.  While detailed content descriptions can be readily supported by large language and multimodal models,
the text descriptions for artistic styles are currently limited to terms like art forms (oil painting, pencil drawing), artists (van Gogh, Picasso)
or simple subjective words (bright, colorful).  Detailed descriptions like color distribution or brush stroke characteristic are not supported yet.
Moreover, for models trained on large amount of uncurated text-image pairs, the probability distribution of styles for any text input is biased towards certain subset reflecting the corresponding bias in training data.
For example, the generated results in the first row of Fig.~\ref{fig:husky} concentrate on a color distribution of black and white, a bias associated with the key word \textit{husky}.  
One main goal of our work is to alleviate these limitations in style specification and diversity using style-guided generation at inference time without the need to retrain
the text-to-image diffusion model.

More technically, diffusion models are powerful in generating realistic and diverse images using an iterative process: given a noisy input $x_t$, estimating and sampling a less noisy
output of $x_{t-1}$ according to the distributions below:
 \begin{equation}
\resizebox{\width}{!}{$
\begin{split}
 x_{t-1} & \sim \mathcal{N}(\mu, \Sigma) \\
 \mu, \Sigma & = \mu_\theta(x_t, t), \Sigma_\theta(x_t, t)
 \end{split}
 $}
 \label{eq:step}
 \end{equation}
\noindent where $\mu_\theta$ and $\Sigma_\theta$ are trained diffusion models that predict the mean and variance of $x_{t-1}$.  Starting from a random noise $x_T$, when the denoising
step is iterated $T$ times, the generated images is denoted as $x_0$.  While this repetitive process is
time consuming, it allows effective fusion of auxiliary information $y$ to steer the sampling process towards a more desired outcome using continuous guidance like
 \begin{equation}
 \resizebox{\width}{!}{$x_{t-1} \sim \mathcal{N}(\mu+g(x_t|y), \Sigma)$}
 \label{eq:guide}
 \end{equation}
\noindent where $g(x_t|y)$ is the guidance function.  In the example of classifier guided
method~\cite{dhariwal_nips_2021}, the sampling process is guided using the gradient
of a classifier which improves the conditional generation of images from a specific class $y$.

Here we propose the first known method to use a style reference image $y$ as a guidance for text-to-image generation.
As $x_t$ is noisy, the direct guidance from the noise-free $y$ would be interfered by the noise.
For the classifier guidance, it is shown that retraining the classifier on noisy data can improve
the guidance quality effectively.
In this work, we propose to move the guidance from the noisy image $x_t$ to $x^t_0$ which is a direct estimate of $x_0$ from $x_t$.
It works effectively without the need of any retraining. As
shown in the second row of Fig.~\ref{fig:husky}, using van Gogh's \textit{The Starry Night} image as a style reference, the proposed style guided generation is able to generate images of
the desired style, accomplishing the functions of both image generation and style transfer in one step.
We have also proposed self style guidance methods without using a style reference.
As shown at the bottom row, it produces a much broader range of styles comparing to the unguided ones at the top.

In summary, we propose an innovative style guided generation method that can enhance existing text-to-image diffusion models for generation
of specific artistic styles or more diverse randomly created styles.  The style guidance function is optimized to minimize the impact of noisy input and maximize the guidance
efficiency.  It is shown that both supervised and self style guidance are effective in generating images of desired styles
while maintaining high relevance of generated images with respect to text inputs.  We have conducted extensive experiments to demonstrate the effectiveness of style guidance on a broad range of potential applications,
including: 1) generating images of a specific style; 2) generating samples from a group of text inputs to
create a series of artwork of the same style; 3) generating from one text input with enhanced style diversity.

\section{Related Works}
\label{sec:rwork}

\subsection{Denoising Diffusion Models}
The latest denoising diffusion models are inspired by non-equilibrium thermodynamics~\cite{sohl_icml_2015}.
They define a Markov chain of diffusion steps to slowly add random noise to data so the intractable real data distribution
is transformed to a tractable one like Gaussian.  Then the models learn to reverse the diffusion process to construct desired data samples from
randomly sample Gaussian noise.  Ho \etal~\cite{ho_nips_2020} proposed a denoising diffusion probabilistic model (DDPM) to interpret
the reverse diffusion process as a large amount of consecutive denoising steps.  For each denoising step, the intermediate output is modeled
as a Gaussian distribution conditional on the input and its mean can be estimated at inference time after properly trained while using
pre-determined variance schedule.
Alternatively, Song \etal~\cite{song_nips_2019, song_iclr_2020} used stochastic differential equations to model the reverse diffusion process
and developed a score-based generative model to produce samples via Langevin dynamics using estimated gradients of the data distribution.
Later numerous methods~\cite{nichol_icml_2021, song_iclr_2021, lu_arxiv_2022} have been proposed to use much fewer denoising steps
without significant degradation in image quality.
While repetitive denoising steps lead to slow sampling time,
it enables flexibility to guide the sampling process for improved image generation quality.
Dhariwal \etal~\cite{dhariwal_nips_2021} proposed a classifier guidance method to iteratively modify the estimated mean according to a gradient
calculated from a classifier retrained with noisy images.
Later Ho \etal~\cite{ho_nipsw_2021} invented a classifier-free guidance method
that trains a conditional model using randomly masked class labels and treat the difference between conditional and unconditional sampling at
inference time as a proxy classifier.  Besides class labels, other auxiliary data can also be used as inference time guidance too.
Choi \etal~\cite{choi_iccv_2021} proposed to use low-resolution images as a guidance to modify the generation process to pull the samplers
towards the reference image iteratively.  Our proposed method is the first known to us that uses image style related features as an inference time
guidance.

\subsection{Text-to-Image Generation}
In recent years, GAN based deep learning models have been successful used for various
generative tasks~\cite{donahue_arxiv_2018, tulyakov_cvpr_2018, karras_cvpr_2019},
including text-to-image generations~\cite{reed_icml_2016, zhang_iccv_2017, xu_cvpr_2018, qiao_cvpr_2019, tao_arxiv_2020, frolov_nn_2021}.
More recently, autoregressive (AR) models
have also shown promising results in image generation~\cite{parmar_icml_2018, chen_icml_2020, esser_cvpr_2021}.  For text-to-image
generations, various frameworks, including DALL-E~\cite{ramesh_icml_2021}, CogView~\cite{ding_arxiv_2021} and M6~\cite{lin_arxiv_2021},
have been proposed to use large transformer structure to model the joint distribution of text and image tokens.
While they have advanced the quality of text-to-image generation greatly,
they are still limited by the weakness of AR models, including unidirectional bias and accumulated prediction errors.
Most recently, diffusion models have shown the capability to push the limit of unconditional image generation.  Consequently, diffusion-based
text-to-image generation has been a red hot research topic in both the academia and industry.


Radford \etal~\cite{radford_icml_2021} first introduced CLIP to learn joint representations between text and images, training an image encoder
and a caption encoder jointly to maximize the dot-product value between the text-image pair.
As CLIP provides a similarity score between an image and a caption, it 
has been used to steer earlier generative models like GANs
to match a user-defined text caption~\cite{gal_arxiv_2021, galatolo_arxiv_2021}.
It was also applied to unconditional diffusion models~\cite{dd_github_2002} as sampling guidance,
showing impressive text-to-image generation capability.
Alternatively, Nichol \etal~\cite{nichol_arxiv_2022} trained a conditional diffusion model (GLIDE) using text-image pairs where the text,
after embedded using a transformer, was used as a conditional input.
Later, Ramesh \etal~\cite{ramesh_arxiv_2022} proposed to use pretrained CLIP image embedding as input for the conditional
image generation model.  For text-to-image generation, a diffusion prior is also trained to generate an image embeddings from the input text CLIP embedding.
Most recently, Saharia \etal~\cite{saharia_arxiv_2022} found that
text embeddings from large language models pretrained on text-only corpora can be used as
remarkably effective conditions for text-to-image synthesis.

\subsection{Arbitrary Style Transfer}
Neural style transfer (NST) refers to a type of methods that transform a digital image to preserve its content
while adopting the visual style of another image.  
Gatys \etal~\cite{gatys_cvpr_2016} defined the style of an image to be multi-level feature correlations (i.e., Gram matrix)
of a trained image classification neural network
and applied style transfer as an iterative optimization problem to balances the content
similarity and style affinity.  To avoid learning for each new style,
more methods~\cite{huang_iccv_2017, li_nips_2017, li_cvpr_2019, park_cvpr_2019, liu_iccv_2021}
are developed to train one model that can transfer an image to any arbitrary artistic style.
Huang \etal~\cite{huang_iccv_2017} first proposed to adjust channel-wise statistics of the content features by adaptive
instance normalization (AdaIN) so that one feature decoder could be trained to generate style-transferred output
using combined scale-adapted content and style losses.
Later Park \etal~\cite{park_cvpr_2019} adopted attention mechanism to match local features of content and style images
and Liu \etal~\cite{liu_iccv_2021} proposed to take both shallow and deep features into account for attention application.
Alternatively Li \etal~\cite{li_nips_2017} replaced adaptive normalization with recursively applied whitening and coloring transformation (WCT)
between the features of the content and style images,
while Li \etal~\cite{li_cvpr_2019} proposed to learn a linear transformation matrix based on arbitrary pairs of content and style images.
Most recently, new style transfer methods~\cite{kwon_cvpr_2022, fu_eccv_2022}
were proposed to define styles using text inputs in replace of style reference images.
Although these methods can be applied to text-to-image generation models after the images are generated,
our proposed method is the first one known to us that can generate images of arbitrary artistic style in one generation process
while maintaining the matching quality between the text and image pair.

\section{Proposed Method}
\label{sec:method}

\subsection{Diffusion Model Background}
Here we adopt the denoising diffusion models introduced by Sohl \etal ~\cite{sohl_icml_2015} and later improved and validated by Ho \etal~\cite{ho_nips_2020} in the more recent DDPM work.
For an image $x_0$ sampled from a distribution $q(x_0)$, a Markov chain of latent variables $x_1, ..., x_T$ can be produced by diffusing the sample using progressively added Gaussian noises:
 \begin{equation}
 \resizebox{\width}{!}{$q(x_t | x_{t-1}) = \mathcal{N}(x_t; \sqrt{1-\beta_t} x_{t-1}, \beta_t \mathrm{I})$}.
 \end{equation}
For each reverse denoising step, 
when the magnitude of the added noise $\beta_t$ is small enough at each step $t$, the posterior $q(x_{t-1} | x_{t})$ can be sufficiently approximated by a diagonal Gaussian.
Additionally, if the magnitude of the total noise added throughout the chain, $1-\bar{\alpha}_T$, is large enough, $x_T$ is well approximated by $\mathcal{N}(0, \mathrm{I})$. Here $\bar{\alpha}_T$ is defined as $\prod_{t=1}^T (1-\beta_t)$.
Based on these approximations, a diffusion model $p_\theta(x_{t-1} | x_t)$ is designed to match the true posterior:
 \begin{equation}
 \resizebox{\width}{!}{$p_{\theta}(x_{t-1}|x_t) = \mathcal{N}(\mu_{\theta}(x_t, t), \Sigma_{\theta}(x_t, t))$}.
 \end{equation}
Starting from a noise $x_T \sim \mathcal{N}(0, \mathrm{I})$, the learned posterior can
be used to sample $x_t$, $t=T-1, T-2, ...$ progressively, resulting in a sampled image $x_0 \sim p_\theta(x_0)$ at the end.

As shown in DDPM, a re-weighted variational lower-bound (VLB) is used as an effective surrogate objective for diffusion model optimization.
Then a diffusion model $\epsilon_{\theta}$ can be trained to predict the added noise using synthesized samples $x_t \sim q(x_t | x_0)$ where a known Gaussian noise $\epsilon$ is added to $x_0$.
This model can then be optimized using a simple standard mean-squared error (MSE) loss:
 \begin{equation}
 \resizebox{\width}{!}{$L^{\text{simple}} = E_{t, x_0, \epsilon} ||\epsilon - \epsilon_{\theta}(x_t, t)||^2$}.
 \end{equation}

This is equivalent to the diffusion model which estimates $\mu_{\theta}$ and $\Sigma_{\theta}$ since $\mu_{\theta}(x_t, t)$ can be derived as
 \begin{equation}
 \resizebox{\width}{!}{$\mu_{\theta}(x_t, t) = \dfrac{1}{\sqrt{1-\beta_t}} \left( x_t - \dfrac{\beta_t}{\sqrt{1-\bar{\alpha}}_t}\epsilon_{\theta}(x_t, t) \right)$}
 \end{equation}
\noindent while $\Sigma_{\theta}$ is  set as a constant.  It is also equivalent to the previous denoising score-matching based models \cite{song_nips_2019, song_iclr_2020}, with the score function $\nabla_{x_t} \log p(x_t) \propto \epsilon_{\theta}(x_t, t)$.  Later Nichol \etal~\cite{nichol_icml_2021} presented a strategy for learning~$\Sigma_{\theta}$, which enables the model to produce high quality samples with fewer diffusion steps.
This learned $\Sigma_{\theta}$ technique is adopted by OpenAI's text-to-image model GLIDE~\cite{nichol_arxiv_2022}, the baseline model used in this work for experiments.

In a follow up work, Dhariwal \etal~\cite{dhariwal_nips_2021} found that even for class-conditional diffusion models, randomly generated samples can be further improved with classifier guidance
at inference time. For the diffusion model with mean~$\mu_{\theta}(x_t,t|y)$ and variance $\Sigma_{\theta}(x_t,t|y)$ where $y$ is the class label, the estimated mean
is perturbed by adding the gradient of the log-probability~$\log p_{\phi}(y|x_t)$ of a target class~$y$ predicted by a classifier.
The resulting new perturbed mean~$\hat{\mu}_{\theta}(x_t, t|y)$ is given by
 \begin{equation}
 \resizebox{\width}{!}{$\hat{\mu}_{\theta}(x_t, t|y) = \mu_{\theta}(x_t,t|y) + s \Sigma_{\theta}(x_t,t|y) \nabla_{x_t} \log p_{\phi}(y|x_t)$}
 \end{equation}
where coefficient $s$ is called the guidance scale.  A larger $s$ leads to higher sample quality but less diversity.

For image-to-text models like GLIDE, similar guidance techniques can be applied by replacing the classifier with a CLIP model.
In this case, the estimated mean during the reverse-process is perturbed by the gradient of the dot product of the paired image and text embeddings:
 \begin{equation}
 \resizebox{\width}{!}{$\hat{\mu}_{\theta}(x_t,t|c) = \mu_{\theta}(x_t,t|c) + s \Sigma_{\theta}(x_t,t|c) \nabla_{x_t} \left(f(x_t) \cdot g(c) \right)$}
 \end{equation}
where $c$ stands for the text input.
Although it is shown~\cite{dd_github_2002} that pretrained CLIP models can be used to guide diffusion models without retuning,
it is better to retrain CLIP on noisy images to obtain the correct gradient in the reverse process.  While our proposed style guidance is inspired by these two guidance
techniques, it is different from them in two major aspects: there is no need for retraining using noisy images (Guide 1) and our guidance scale adaptive (Guide 2).

\subsection{Supervised Style Guidance}
The motivation of style guided diffusion is to generate images with desired styles.
Following the examples of classifier and CLIP guidance, we can design a simple style guidance method as
 \begin{equation}
 \resizebox{\width}{!}{$\hat{\mu}_{\theta}(x_t, t) = \mu_{\theta}(x_t,t) - s \cdot \Sigma_{\theta}(x_t,t) \nabla_{x_t} \left|f(x_t)-f(y) \right|$}
 \end{equation}
where $f$ is the style feature function and $y$ is the style reference image. The style distance needs to be subtracted as the guidance because the aim is to minimize style differences, as opposed to adding classifier guidance in the case of class guidance to maximize class probability.
It is not clear if existing style feature function $f$ is robust to noisy images.  We propose the two guidance techniques, Guide 1 and Guide 2 mentioned above, to mitigate this uncertainty.

Guide 1. In other works, the perturbing gradient is calculated by comparing a noisy image with a reference like class label
or text, which is trained from noise-free images.  Thus retraining the associated classifier or CLIP model with noisy images is helpful.
In the case of style guidance though, there is overlap between image noises and certain style characteristics.
To avoid this confusion and additional training, here we propose an alternative guidance method to calculate
the perturbing gradient by comparing the "noise-free" $x^t_0$ and reference $y$ instead
 \begin{equation}
 \resizebox{\width}{!}{$
 \begin{split}
 x^t_0 & = (x_t-\sqrt{1-\bar{\alpha}_t}\epsilon_{\theta}(x_t, t)) / \sqrt{\bar{\alpha}_t} \\
 \hat{\mu}_{\theta}(x_t, t) & = \mu_{\theta}(x_t,t) - s \Sigma_{\theta}(x_t,t) \nabla_{x_t} \left|f(x^t_0)\!-\!f(y) \right|
 \end{split}
 $}
 \label{eq:sg}
 \end{equation}

Guide 2. As $x^t_0$ is only noise-free theoretically because the noise estimation can not be perfect
in one step, we have found empirically that it is more effective to increase guidance scale
when the noise level is lower.  As a result $s$ is set as an adaptive variable here as
 \begin{equation}
 \resizebox{\width}{!}{$s = s_0/\sqrt{\Sigma_{\theta}(x_t,t)}$}
 \end{equation}
where $s_0$ is a constant denoted as the base scale.

\subsection{Style Features}
For style features used to guide the reverse-diffusion process, we adopt the instance normalization (IN) statistics
used in AdaIN~\cite{huang_cvpr_2017} for its cleanness over the original Gramm matrix features.
For an image $x$, its style features $f(x)$ is defined as
 \begin{equation}
 \resizebox{\width}{!}{$f(x) = \{ \lambda_i \eta(\psi_i(x)), \lambda_i \sigma(\psi_i(x)) \mid i \in [1,4]\}$}
 \end{equation}
where $\psi_i$ denotes a layer in VGG-19, $\eta$ and $\sigma$ represent the mean and standard deviation respectively, and $\lambda_i$ is the weight of layer $i$.
We use $\texttt{relu1\_1}$, $\texttt{relu2\_1}$, $\texttt{relu3\_1}$, $\texttt{relu4\_1}$ layers for style feature calculation
and use equal weights for style loss assessment following the previous practices.  But for style guidance, optimal weights for each layer are selected
empirically for the best guidance effects.  Additionally, while the standard MSE loss is used as style loss in result assessment,
the mean absolute error (MAE) loss is found to be more effective when used to calculate the perturbing gradient during style guidance.

\subsection{Self Style Guidance}
For current text-to-image generation models, as the sample is randomly generated,
it often takes multiple samples for one text input to achieve high image quality
and text-image relevance.  As a batch generation from one text input is already
required, we propose a self style guidance method to sample a more diverse styles
within the batch, breaking the limitation of biased style associated with
a given type of object as shown in Fig.~\ref{fig:husky} earlier.
It follows the same style guidance principle as the supervised
one but does not require a style reference, hence self guidance.
Mathematically, the guidance correction is defined as 
 \begin{equation}
 \resizebox{\width}{!}{$\mu_{\theta}(x_t,t) + s \cdot \Sigma_{\theta}(x_t,t) \nabla_{x_t} \nu_f(x^t_0)$}
 \end{equation}
where $\nu_f$ represents the variance in style features $f$.
Here we denote it as \textbf{contrastive self guidance} since it aims to increase the style contrasts within the batch
without using a style reference.

 \begin{figure*}[t]
 \captionsetup[subfigure]{labelformat=empty}
 \begin{center}
  \begin{subfigure}[b]{\textwidth}
    \centering
     \includegraphics[width=0.12\textwidth, interpolate=false]{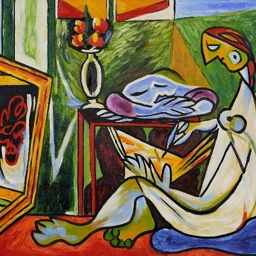} \hspace*{-3pt} \includegraphics[width=0.72\textwidth, interpolate=false]{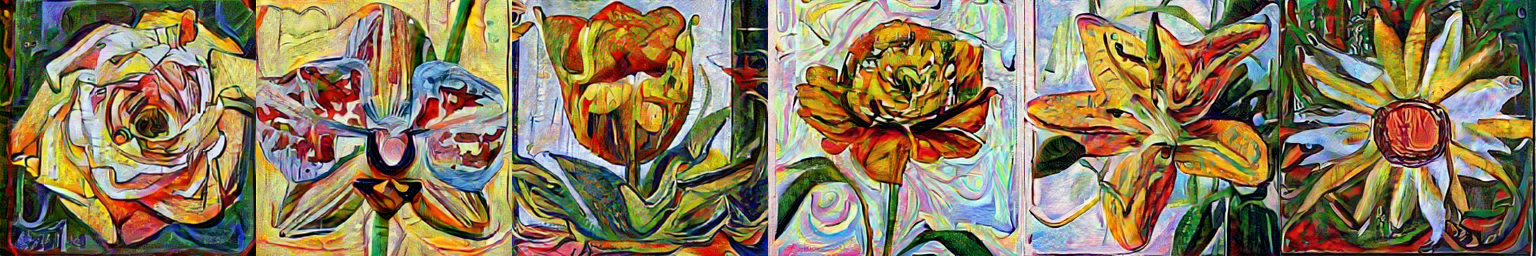} \hspace*{0pt}
     \begin{minipage}{0.14\textwidth}\vskip-60pt \scriptsize{\textit{\blue{an oil painting of a/an} rose; orchid; tulip; peony; lily; daisy flower}}\end{minipage}
     \vspace{0pt}
     \includegraphics[width=0.12\textwidth, interpolate=false]{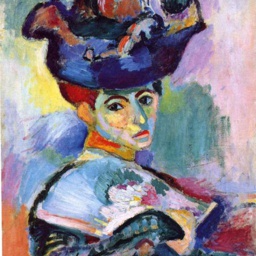} \hspace*{-3pt} \includegraphics[width=0.72\textwidth, interpolate=false]{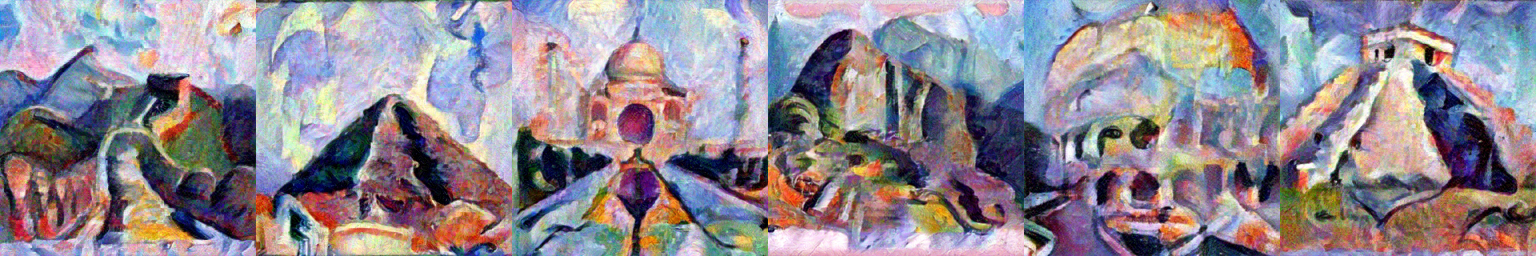} \hspace*{0pt}
     \begin{minipage}{0.14\textwidth}\vskip-60pt \scriptsize{\textit{the Great Wall; Great Pyramid of Giza; Taj Mahal; Machu Picchu; Colosseum, Rome;Chichen Itza, Mexico}}\end{minipage}
     \vspace{0pt}
     \includegraphics[width=0.12\textwidth, interpolate=false]{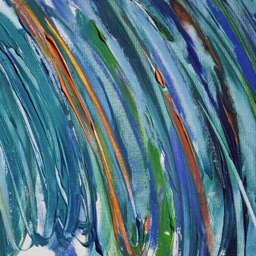} \hspace*{-3pt} \includegraphics[width=0.72\textwidth, interpolate=false]{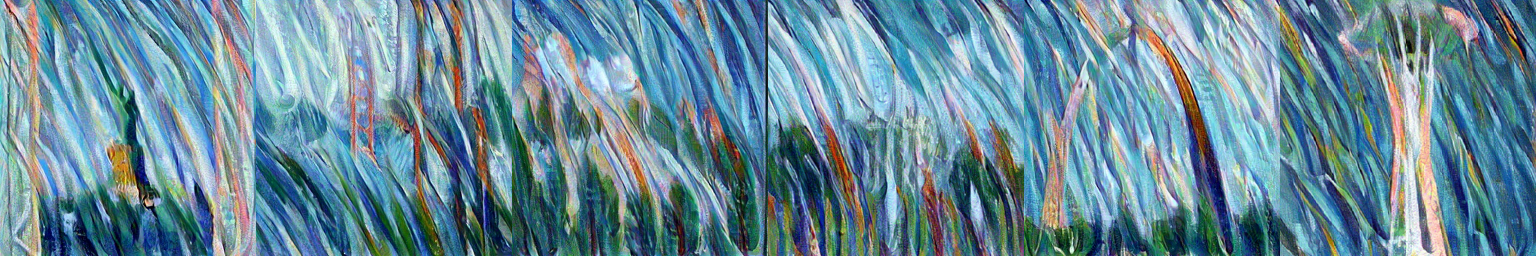} \hspace*{0pt}
     \begin{minipage}{0.14\textwidth}\vskip-60pt \scriptsize{\textit{the Statue of Liberty; Golden Gate Bridge; Mount Rushmore National Memorial; White House; Gateway Arch; Space Needle}}\end{minipage}
     \vspace{0pt}
     \includegraphics[width=0.12\textwidth, interpolate=false]{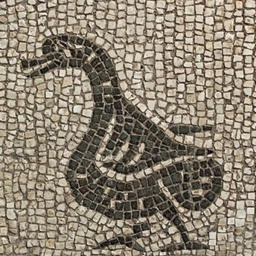} \hspace*{-3pt} \includegraphics[width=0.72\textwidth, interpolate=false]{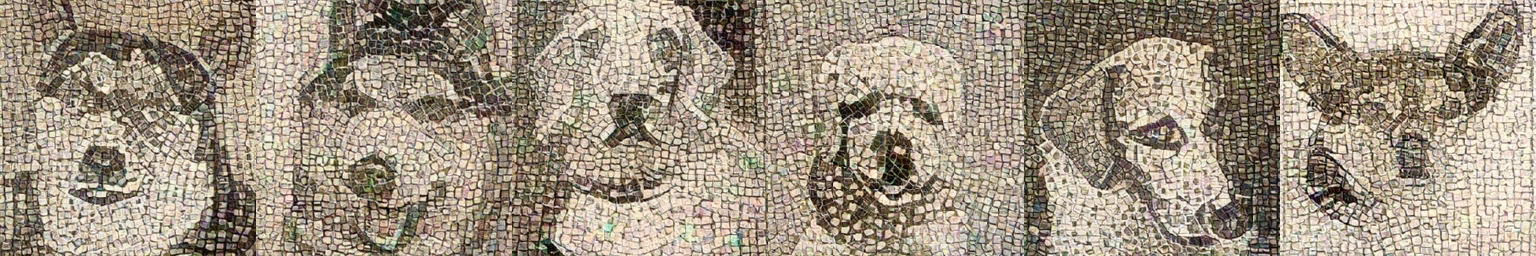} \hspace*{0pt}
     \begin{minipage}{0.14\textwidth}\vskip-60pt \scriptsize{\textit{corgi; husky; golden retriever; poodle; beagle; chihuahua}}\end{minipage}
     \vspace{0pt}
     \hspace*{0.12\textwidth} \includegraphics[width=0.72\textwidth, interpolate=false]{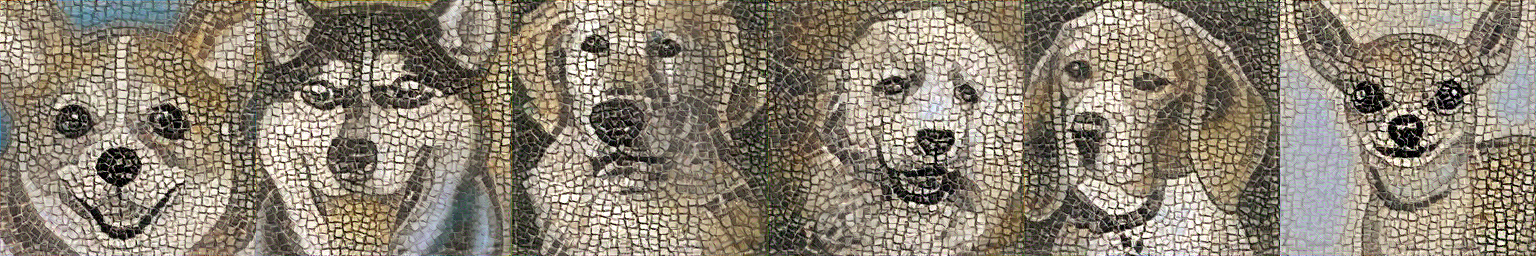} \hspace*{0pt}
     \begin{minipage}{0.14\textwidth}\vskip-60pt \small{Style transfer after unguided sampling using Gatys~\cite{gatys_cvpr_2016}}\end{minipage}
     \vspace{0pt}
     \hspace*{0.12\textwidth} \includegraphics[width=0.72\textwidth, interpolate=false]{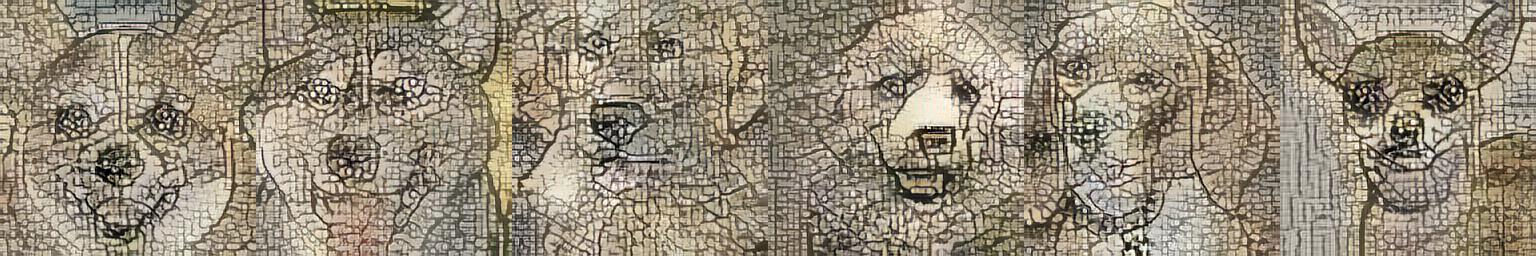} \hspace*{0pt}
     \begin{minipage}{0.14\textwidth}\vskip-60pt \small{Style transfer after unguided sampling using AdaIN~\cite{huang_iccv_2017}}\end{minipage}
  \end{subfigure}
 \end{center}
 \vspace{-9pt}
 \caption{Visual examples generated from supervised style guidance with diverse styles and various image generation text inputs (style reference on the left).  Results from two-step generations are included for comparison}
 \vspace{-9pt}
 \label{fig:sg}
 \end{figure*}

Alternatively, for artwork creation, designing a series of work using the same artistic style is often needed, like for stamps or posters.  For this application,
we propose a \textbf{synonymous self guidance} method to generate multiple images in one shared style from a set of text inputs, again
without using a style reference.  To increase the style diversity in this method,
a mixed style feature is first proposed for a set of images $x$, defined as
 \begin{equation}
 \resizebox{\width}{!}{$f_m(x) = \{ \lambda_i \eta(\psi_i(x_{r_i})), \lambda_i \sigma(\psi_i(x_{r_i})) \mid i \in [1,4]\}$}
 \end{equation}
where $r_i$ is a random index number to associate features from layer $i$ with one image $x_{r_i}$
during each sampling.  The synonymous self guidance is then applied as
 \begin{equation}
 \resizebox{\width}{!}{$\mu_{\theta}(x_t,t) - s \cdot \Sigma_{\theta}(x_t,t) \nabla_{x_t} \left|f(x^t_0)-f_m(x^t_0) \right|$}.
 \end{equation}
Note that $f_m(x^t_0)$ is a dynamic style reference which changes at each denoising step $t$, enabling
the creation of more diverse styles from iterative sampling of dynamic features mixed from multiple
images.

\section{Experiments}
\label{sec:exp}
All experiments in this study, unless mentioned otherwise, are conducted using OpenAI's GLIDE model~\footnotemark[\value{footnote}].
While our method applies to different versions, we evaluate our method based on the public filtered version with image size $256 \times 256$.
So the results are only comparable to images generated by this model in fidelity and text-image similarity.
Additionally, as there is no reference image set with the same content and style distribution as
our generated sets, image quality metrics like FID~\cite{heusel_nips_2017} which need a ground-truth reference are not applicable.
On the other hand, the CLIP score is applicable here.
It is defined as correlation between the CLIP text embedding and image embedding and can be used to assess text-image similarity under style guidance.
Besides, unlike other models, the GLIDE version used here does not use CLIP guidance for generation, avoiding
impact on fairness of the CLIP score metric.
The specific CLIP model used for testing here is ViT-B/32.

To investigate the performance of supervised style guidance with a style reference image, we selected 12 random
artworks from WikiArt~\cite{saleh_arxiv_2015} as in AdaIN~\cite{huang_iccv_2017}.  For text inputs, we organized them in 5 different content categories,
including dogs, flowers, wonders of the world, American landmarks and general places like parks.  For each category, there are 6 specific inputs.  For dogs,
they are "\textit{an oil painting of a/an corgi/husky/golden retriever/poodle/beagle/chihuahua}".  These are chosen to demonstrate the application
of creating a group of artworks with similar types of contents while using the same artistic style, just like designing a set of stamps or posters.

 \begin{figure}[t]
 \begin{center}
     \includegraphics[width=0.65\linewidth]{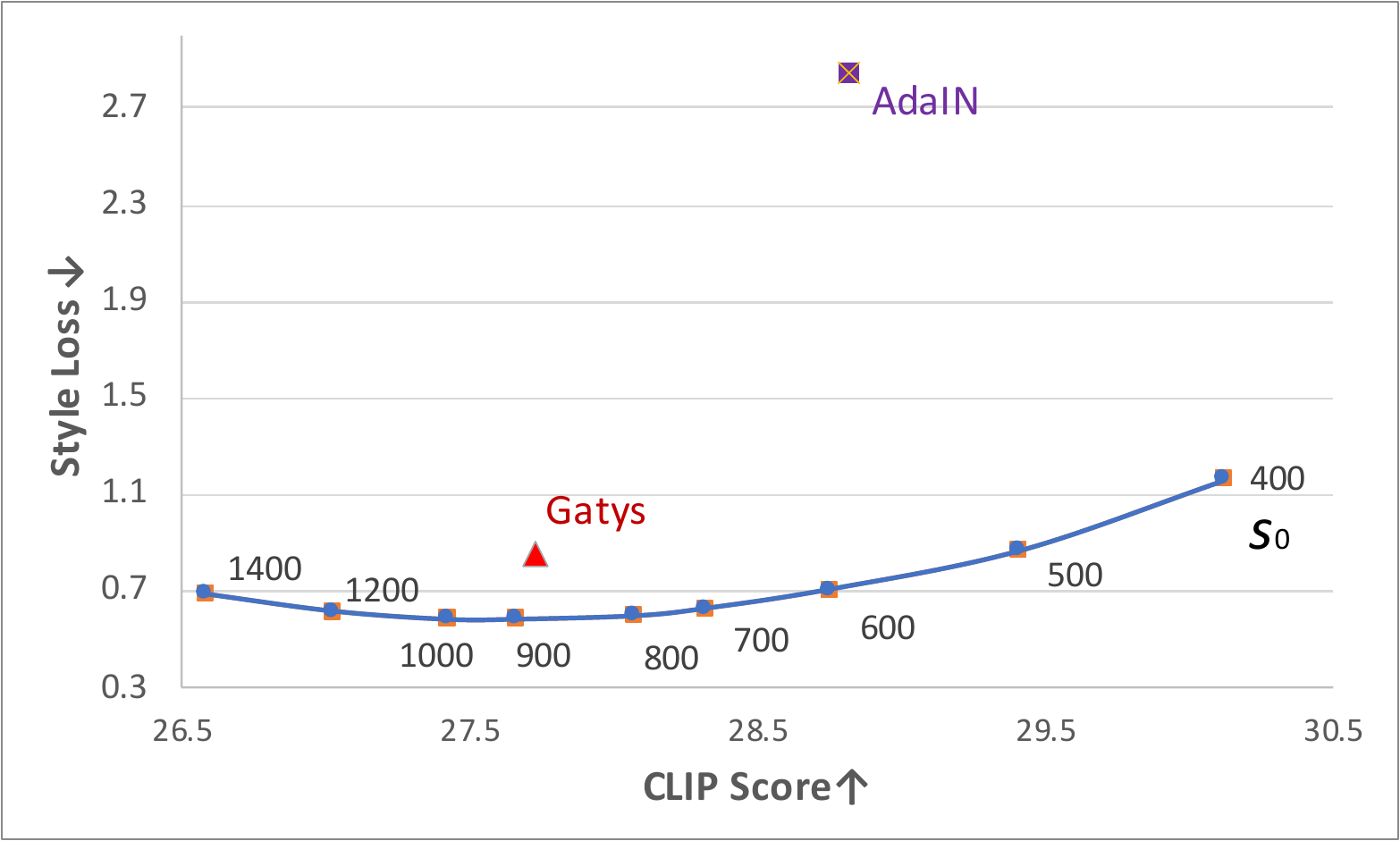}
 \end{center}
 \vspace{-6pt}
 \caption{Trade-off curve for supervised style guidance with varying $s_0$ values.  Two-step results from style transfer applied to unguided samples are included for comparison.}
 \vspace{-3pt}
 \label{fig:slcs}
 \end{figure}

\subsection{Supervised Style Guidance}
As shown in Fig.~\ref{fig:sg}, our proposed style guidance method is able to generate images for a range of subjects while following an arbitrary style from a reference image
at the same time.  For the guided samples in the first four rows, each set of images are generated using the style reference
on the left and the set of text inputs on the right.  Each set are generated in one sampling process, instead of selection
from multiple sampling processes.
For the challenging case in the third row where the style reference consists of only simple color strokes without semantic information,
our style guidance method is still effective in creating relevant contents, like the unique architecture pattern
in the center for the example of the \textit{White House}.
For the fourth example, additional results from two-step generation methods, applying NST after unguided
sample, are also included for comparison.
Two style transfer methods, Gatys~\cite{gatys_cvpr_2016} and AdaIN~\cite{huang_iccv_2017},
are applied to an unguided sample using the same style reference as the guided sample above them.
It shows that our one-step style guided result have higher consistency with
the style reference while the two-step ones have some residual color artifacts, like blur background in the first Gatys image.
Quantitative assessments of these three methods are included in Fig.~\ref{fig:slcs}.

 \begin{figure*}[t]
 \captionsetup[subfigure]{labelformat=empty}
 \begin{center}
  \begin{subfigure}[b]{\textwidth}
    \centering
     \includegraphics[width=0.8\textwidth, interpolate=false]{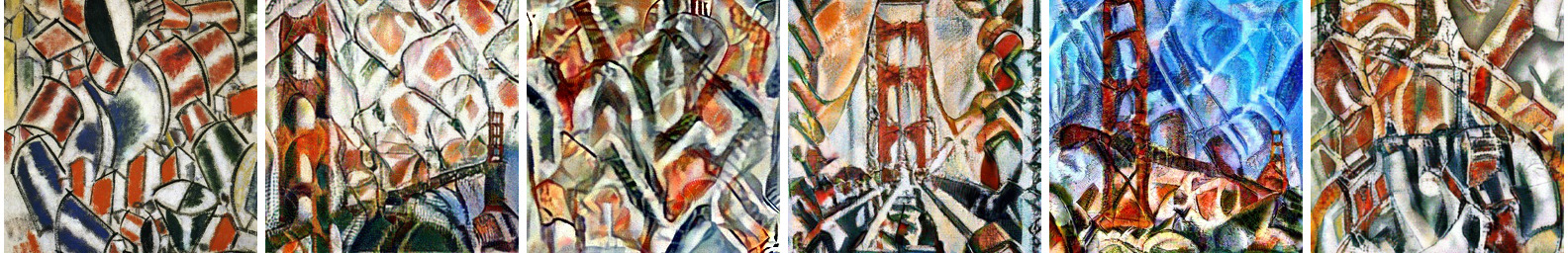} \hspace*{3pt}
     \begin{minipage}{0.15\textwidth}\vskip-60pt \small{\textit{an oil painting of the Golden Gate Bridge}}\end{minipage}
     \caption{\hspace{-86pt}Style Image \hspace{15pt} \#0: Optimal \hspace{19pt} \#1: $(x_t, y)$ \hspace{15pt} \#2: Fixed Scale \hspace{4pt} \#3: Equal Weights \hspace{11pt} \#4: MSE}
  \end{subfigure}
 \end{center}
 \vspace{-12pt}
 \caption{Visual examples of different style guidance settings, demonstrating degradations in either style loss or text-image similarity in alternative settings in comparison to the optimal \#0.}
 \vspace{-3pt}
 \label{fig:settings}
 \end{figure*}

 \begin{figure*}[t]
 \captionsetup[subfigure]{labelformat=empty}
 \begin{center}
  \begin{subfigure}[b]{\textwidth}
    \centering
     \includegraphics[width=0.72\textwidth, interpolate=false]{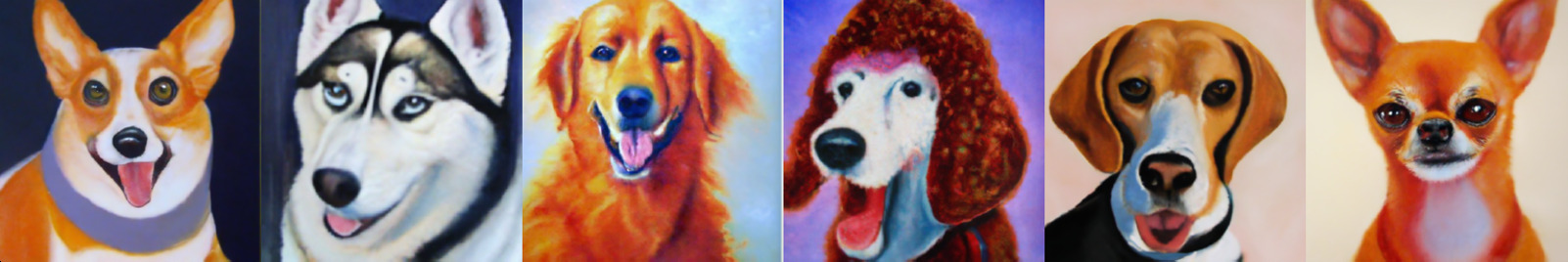} \hspace*{3pt}
     \begin{minipage}{0.2\textwidth}\vskip-55pt \footnotesize{\textbf{Unguided Sampling}:} \newline \footnotesize{\textit{\blue{an oil painting of a} \red{happy} corgi; husky; golden retriever; poodle; beagle; chihuahua}}\end{minipage}
     \vspace{0pt}
     \includegraphics[width=0.72\textwidth, interpolate=false]{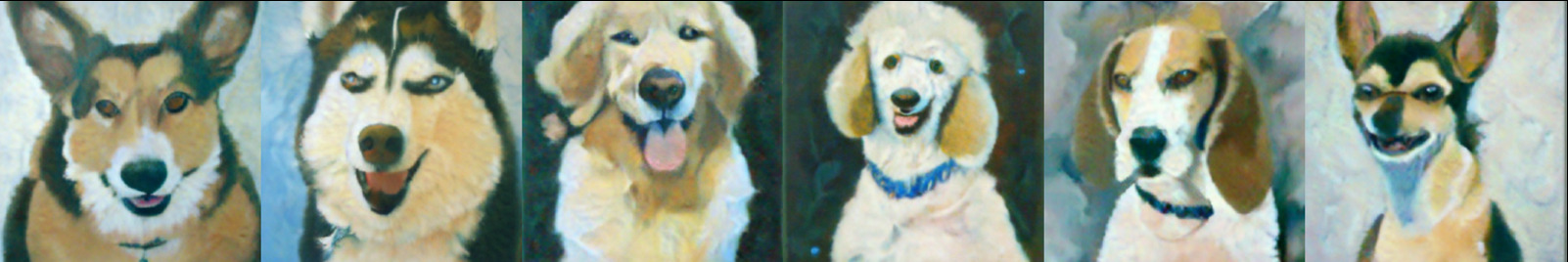} \hspace*{3pt}
     \begin{minipage}{0.2\textwidth}\vskip-55pt \footnotesize{\textbf{Self Style Guidance}:} \newline \footnotesize{\textit{\blue{an oil painting of a} \red{happy} corgi; husky; golden retriever; poodle; beagle; chihuahua}}\end{minipage}
     \vspace{0pt}
     \includegraphics[width=0.72\textwidth, interpolate=false]{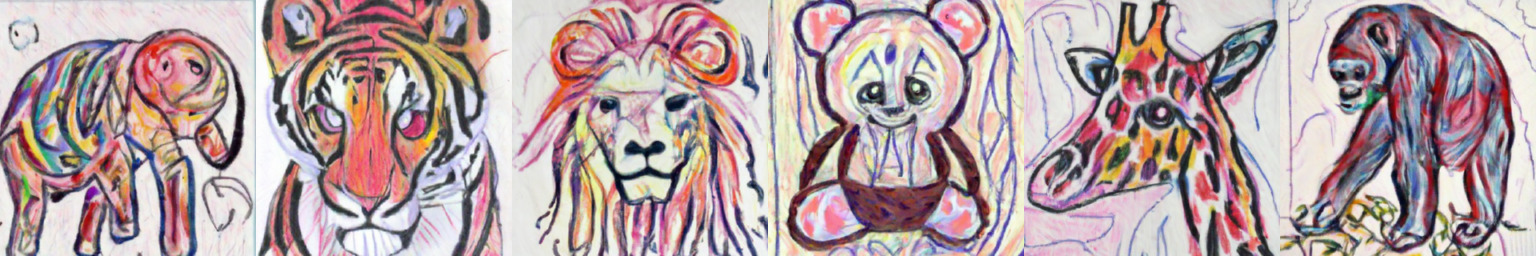} \hspace*{3pt}
     \begin{minipage}{0.2\textwidth}\vskip-60pt \footnotesize{\textit{\blue{a \red{crayon drawing} of a/an} elephant; tiger; lion; panda; giraffe; gorilla}}\end{minipage}
     \vspace{0pt}
     \includegraphics[width=0.72\textwidth, interpolate=false]{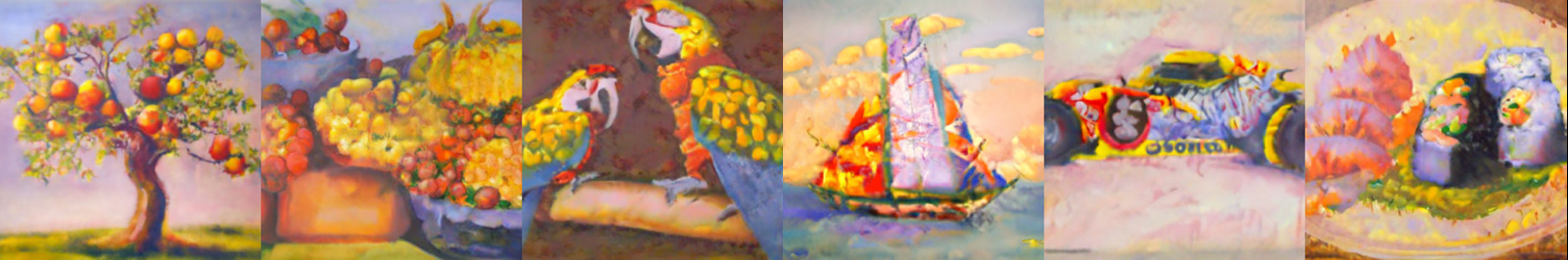} \hspace*{3pt}
     \begin{minipage}{0.2\textwidth}\vskip-60pt \footnotesize{\textit{\blue{an oil painting of a/an} apple tree; Bazaar market; parrot; sailboat; F1 race car; sushi plate}}\end{minipage}
  \end{subfigure}
 \end{center}
 \vspace{-6pt}
 \caption{Visual examples of synonymous self  guidance: create a set of images with one shared style. It produces diverse styles while maintaining content fidelity for variations in text inputs like \textit{\red{happy}} expression and \textit{\red{crayon drawing}}.}
 \vspace{-3pt}
 \label{fig:ssg}
 \end{figure*}

\subsection{Style Guidance Optimization}
\label{sec:sgo}
To maximize the efficiency of style guidance, multiple settings in style features and guidance methods are investigated.
First the trade-off between the style loss and CLIP score is studied by varying the base
guidance scale $s_0$, using all 12 style images and 5 categories of text inputs.
As shown in Fig.~\ref{fig:slcs}, when the guidance scale decreases, it leads to a higher style loss naturally,
which results in a higher CLIP score for better text-image similarity.  However, when the guidance scale increases, while the CLIP score decrease
continuously, the style loss decreases first but increases after reaching the minimum around $s_0 = 1000$.  This reverse trend is similar in nature to the divergence issue
caused by large learning rate, as the guidance scale controls the step size of gradient guidance.
The results from two-step methods, applying style transfer after unguided sampling, are also included in Fig.~\ref{fig:slcs} for comparison.
AdaIN~\cite{huang_iccv_2017} uses a one step decoding process to apply arbitrary style transfer so understandably it
has a higher style loss than Gatys~\cite{gatys_cvpr_2016} which applies an iterative learning process to transform
the image.  Our style guidance is also applied iteratively during the reverse denoising process, similar to Gatys in this aspect.  As there is
not a single image reference of content image for a given text input, the style guided generation is able to achieve lower style loss than Gatys as it can adjust its content
accordingly given the style guidance.


\begin{table}[h!]
	\centering
	\footnotesize
	\setlength{\tabcolsep}{4pt}
\vspace{0pt}
	\caption{Ablation study for style guidance settings. \red{Red} highlights suboptimal settings and resulted degradations.}
\vspace{3pt}
	\begin{tabular}{ccccccc} 
		\hline
		{Setting} & {Guidance} & {Style} & {Adaptive} & {Varying} & {CLIP$\wt{^{^a}}$} & {Style$\wt{^{^a}}$} \\
		{\#$\;\;$} & {Pair} & {Distance} & {Scale} & {Weights} & {Score$\uparrow$} & {Loss$\downarrow$} \\
		\hline \hline
		\#0$\wt{^{^A}}$ & {$(x^t_0, y)$} & {MAE} & {\cmark} & {\cmark} & 27.45 & 0.58 \\
		\#1$\wt{^{^A}}$ & {$(\red{x_t}, y)$} & {MAE} & {\cmark} & {\cmark} & \red{24.39} & 0.59 \\
		\#2$\wt{^{^A}}$ & {$(x^t_0, y)$} & {MAE} & \red{\xmark} & {\cmark} & 27.64 & \red{0.82} \\
		\#3$\wt{^{^A}}$ & {$(x^t_0, y)$} & {MAE} & {\cmark} & \red{\xmark} & 27.72 & \red{1.57} \\
		\#4$\wt{^{^A}}$ & {$(x^t_0, y)$} & \red{MSE} & {\cmark} & {\cmark} & \red{25.73} & 0.48 \\
		\hline	\end{tabular}
\label{tab:ablation}
\vspace{-9pt}
\end{table}

 \begin{figure*}[t]
 \captionsetup[subfigure]{labelformat=empty}
 \begin{center}
  \begin{subfigure}[b]{\textwidth}
    \centering
     \includegraphics[width=0.80\textwidth, interpolate=false]{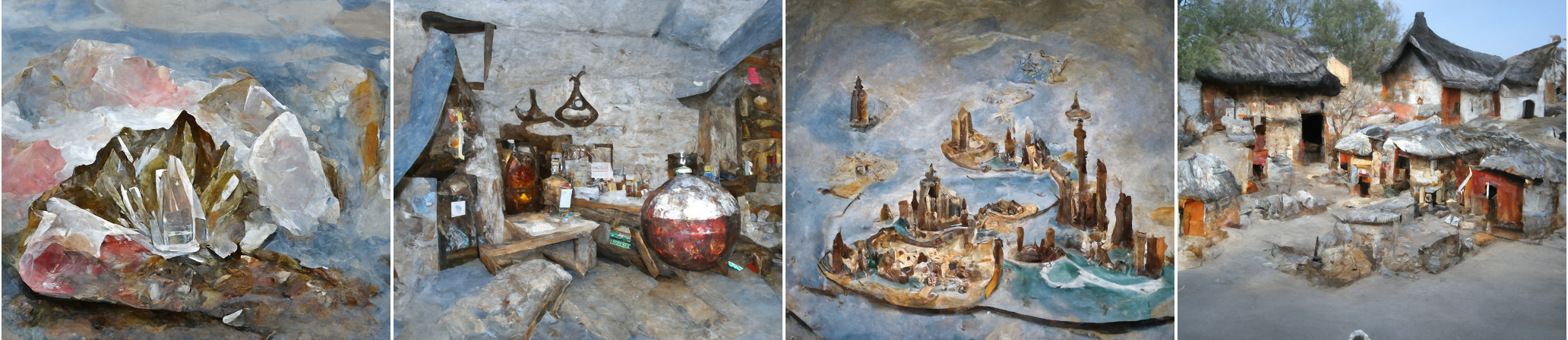} 
     \begin{minipage}{0.18\textwidth}\vskip-70pt \footnotesize{\textit{A beautiful painting of a quartz crystal in a serene landscape;
            Rustic interior of an alchemy shop; A beautiful painting of a map of the city of Atlantis; Ancient Chinese village}}\end{minipage}
  \end{subfigure}
 \end{center}
 \vspace{-6pt}
 \caption{Visual examples generated from Disco Diffusion~\cite{dd_github_2022} using synonymous self guidance.}
 \vspace{-3pt}
 \label{fig:disco}
 \end{figure*}

Secondly, an ablation study is conducted to compare different settings studied for effective style guidance.
For optimal settings, the gradient calculation is conducted between $x^t_0$, in contrast to using $x_t$, and noise-free reference $y$,
MAE is used for calculation of the style feature distance, adaptive guidance scale is used in place of constant scale,
and optimal varying weights are used for different style feature layers.
As shown in Table~\ref{tab:ablation}, results of CLIP score and style loss for the optimal setting are shown in the first row.
For other settings in the following rows, one aspect of the optimal setting is changed.  For each set of setting,
the base scale $s_0$ is adjusted accordingly to get best
overall performance of its own.  It is shown in both Table~\ref{tab:ablation} and Fig.~\ref{fig:settings} that
the optimal set of setting leads to the best overall quality in style fidelity and text-image similarity.
When the perturbing gradient for style guidance is calculated from $x_t$ as
used in previous guidance methods, it has a significantly lower CLIP score, demonstrated
by the unrecognizable objects in Fig.~\ref{fig:settings} (\#1).
For constant guidance scale, the CLIP score is equivalent to the adaptive one but the style loss increases significantly.
Similarly, replacing the customized varying weights with equal ones leads to even worse
performance in style loss, demonstrated as undesirable blue backgrounds in Fig.~\ref{fig:settings} (\#3).
For the distance metric to compare styles of $x^t_0$ and $y$, MSE results in slightly lower style
loss than MAE, but it has a worse CLIP score, often leading to contents not matching the text input, like
in the \textit{Golden Gate Bridge} painting in Fig.~\ref{fig:settings}.

\subsection{Self Style Guidance}
\begin{figure}[t]
 \captionsetup[subfigure]{labelformat=empty}
 \begin{center}
  \begin{subfigure}[b]{0.8\linewidth}
    \centering
      \includegraphics[width=\linewidth, interpolate=false]{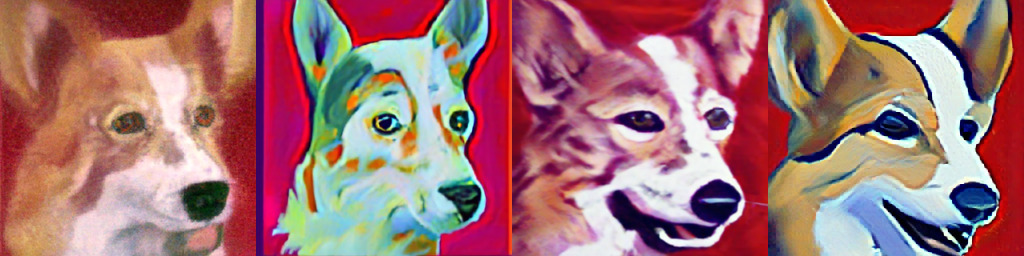}
  \end{subfigure}
 \end{center}
 \vspace{-6pt}
 \caption{Visual examples from contrastive self guidance, demonstrating larger variations in created styles.}
 \vspace{-3pt}
 \label{fig:st}
 \end{figure}

A comprehensive set of experiments are conducted to demonstrate the capability of self style guidance.
For synonymous self guidance, as shown in Fig~\ref{fig:ssg}, it is compared with unguided sampling in
generation a set of images, organized in rows for each set of samples.
The unguided sampling is able to generate realistic images,
the style of each type of object tends to bias towards its natural appearance, like the black and white husky.
In comparison, self guided samples are similar in text-to-image similarity like rendering
the \textit{\red{happy}} expressions faithfully and there are vibrant and diverse styles.
It also shows robustness to different object types and base styles as defined in text inputs.
As shown in the last example, applying synonymous self guidance to a mixed set of objects
may create brand new styles, like the same round patterns appearing as apples, clouds or feathers depending on
object type.  The examples of contrastive style guidance are included in
Fig.~\ref{fig:st}, where an additional constraint on content is also applied
to focus on the increased variance in style.
Lastly, Fig.~\ref{fig:disco} shows that self style guidance is also applicable to
other models like Disco Diffusion~\cite{dd_github_2002},
generating realistic high resolution ($512 \times 448$, resized to $256 \times 224$ due to file size limit)
images from a mixed set of text inputs, sharing the same created style.

\begin{figure}[t]
 \begin{center}
     \includegraphics[width=0.75\linewidth]{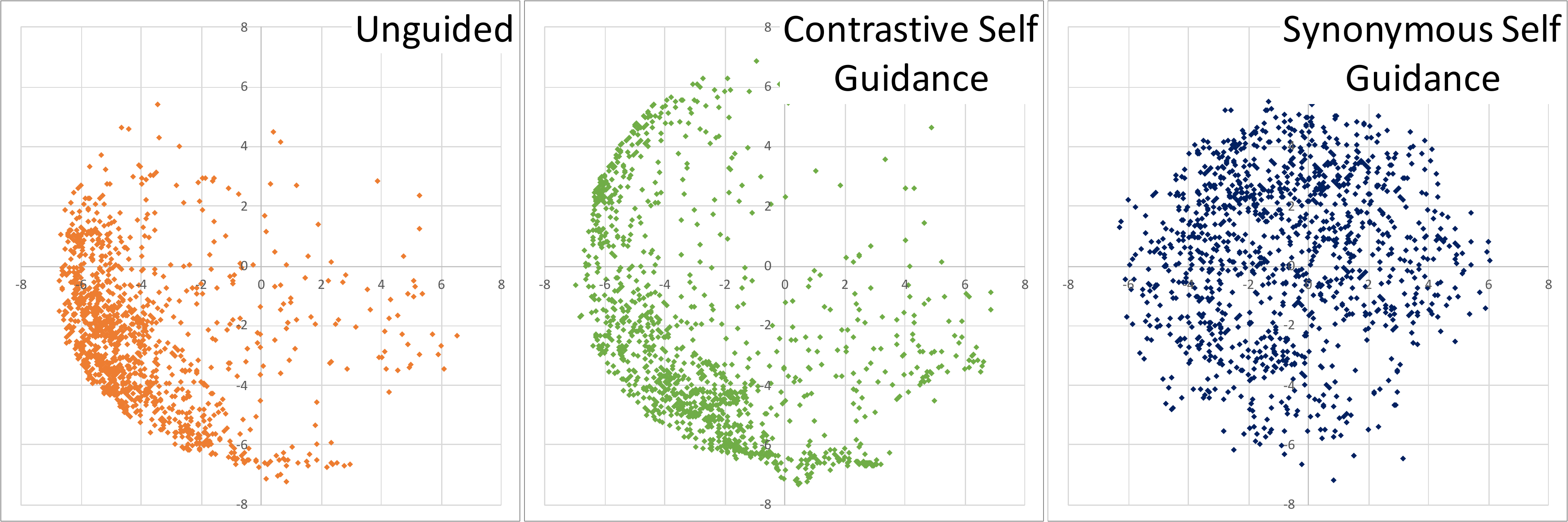}
 \end{center}
 \vspace{-6pt}
 \caption{Style diversity comparison between unguided sampling and self style guided 1200 samples each from "\textit{an oil painting of a husky}", plotted in compressed 2-dimensional space using t-SNE~\cite{van_jmlr_2008}.}
 \vspace{-3pt}
 \label{fig:tsne}
 \end{figure}

Using the same test set as in supervised style guidance, we have compared self style guided sampling with unguided ones in terms of
text-image similarity and style diversity.  For text-image similarity, the average CLIP scores are close to each other at
34.56, 34.12 and 33.56 for unguided, contrastive self guidance
and synonymous self guidance respectively.  Beside,
for synonymous guidance, the average style loss within each generated batch is only
0.27, making it a great tool to generate a set of images with almost identical styles.
For style diversity, as visualized using t-SNE~\cite{van_jmlr_2008} in Fig.~\ref{fig:tsne},
compared to unguided sampling, contrastive self guidance is able to increase variations
in styles but still has dense distribution in some regions.  In comparison,
synonymous self guidance have a near uniform distribution over a large range,
demonstrating that the proposed mixed style reference helps the generation model
sample from styles which are not commonly seen in the training dataset. 

\section{Conclusions}

In this paper, we present a simple and effective style guidance method which helps diffusion-based text-to-image generation models to generate image of desirable artistic styles.
It is applied to inference only, without the need to change other aspects of diffusion models.
Key innovations like applying guidance correction to the "noise-free" $x^t_0$ instead of noisy $x_t$
and adaptive guidance scale and style feature weights are proposed to optimize the effectiveness of style guidance.
For supervised style guidance, it is able to generate images using the style characteristics of a reference image in one step, achieving lower style loss than using additional neural style transfer after unguided sampling.
For self style guidance without a reference, it not only generates realistic images with high text-image similarity, but also creates more diverse styles than unguided sampling.
For synonymous self guidance, it generates multiple images sampled from a set of text inputs in one process,
created with a shared style.  For contrastive self guidance, it increases style diversity in samples generated from the same text input.
The proposed method is validated using a comprehensive set of text inputs, reference styles, guidance options and diffusion models.

The 2015 AdaIN~\cite{huang_iccv_2017} work is used as the main baseline because more recent works focus on model innovation
which is not applicable to our method without model change and additional training.
For applicable style features, they mostly adopt the ones used in AdaIN with minor variations.
An interesting direction for future development is applying more advanced style features like context-aware ones~\cite{liao_tip_2022}.

\bibliographystyle{unsrt}

\begin{thebibliography}{1}

\bibitem{avrahami_cvpr_2022}
Omri Avrahami, Dani Lischinski, and Ohad Fried.
\newblock Blended diffusion for text-driven editing of natural images.
\newblock In {\em Proceedings of the IEEE/CVF Conference on Computer Vision and
  Pattern Recognition}, pages 18208--18218, 2022.

\bibitem{chen_icml_2020}
Mark Chen, Alec Radford, Rewon Child, Jeffrey Wu, Heewoo Jun, David Luan, and
  Ilya Sutskever.
\newblock Generative pretraining from pixels.
\newblock In {\em International conference on machine learning}, pages
  1691--1703, 2020.

\bibitem{choi_iccv_2021}
Jooyoung Choi, Sungwon Kim, Yonghyun Jeong, Youngjune Gwon, and Sungroh Yoon.
\newblock {ILVR}: Conditioning method for denoising diffusion probabilistic
  models.
\newblock In {\em 2021 IEEE/CVF International Conference on Computer Vision
  (ICCV)}, pages 14347--14356. IEEE, 2021.

\bibitem{dd_github_2002}
Katherine Crowson, Maxwell Ingham, Adam Letts, and Alex Spirin.
\newblock Disco {Diffusion}.
\newblock \url{https://github.com/alembics/disco-diffusion}, 2022.

\bibitem{dhariwal_nips_2021}
Prafulla Dhariwal and Alexander Nichol.
\newblock Diffusion models beat {GANs} on image synthesis.
\newblock {\em Advances in Neural Information Processing Systems},
  34:8780--8794, 2021.

\bibitem{ding_arxiv_2021}
Ming Ding, Zhuoyi Yang, Wenyi Hong, Wendi Zheng, Chang Zhou, Da Yin, Junyang
  Lin, Xu Zou, Zhou Shao, Hongxia Yang, et~al.
\newblock {CogView}: Mastering text-to-image generation via transformers.
\newblock {\em Advances in Neural Information Processing Systems},
  34:19822--19835, 2021.

\bibitem{donahue_arxiv_2018}
Chris Donahue, Julian McAuley, and Miller Puckette.
\newblock Adversarial audio synthesis.
\newblock {\em arXiv preprint arXiv:1802.04208}, 2018.

\bibitem{esser_cvpr_2021}
Patrick Esser, Robin Rombach, and Bjorn Ommer.
\newblock Taming transformers for high-resolution image synthesis.
\newblock In {\em Proceedings of the IEEE/CVF conference on computer vision and
  pattern recognition}, pages 12873--12883, 2021.

\bibitem{frolov_nn_2021}
Stanislav Frolov, Tobias Hinz, Federico Raue, J{\"o}rn Hees, and Andreas
  Dengel.
\newblock Adversarial text-to-image synthesis: A review.
\newblock {\em Neural Networks}, 144:187--209, 2021.

\bibitem{fu_eccv_2022}
Tsu-Jui Fu, Xin~Eric Wang, and William~Yang Wang.
\newblock Language-driven artistic style transfer.
\newblock In {\em Proceedings of the European Conference on Computer Vision
  (ECCV)}, 2022.

\bibitem{gal_arxiv_2021}
Rinon Gal, Or Patashnik, Haggai Maron, Gal Chechik, and Daniel Cohen-Or.
\newblock {StyleGAN-NADA}: {CLIP}-guided domain adaptation of image generators.
\newblock {\em arXiv preprint arXiv:2108.00946}, 2021.

\bibitem{galatolo_arxiv_2021}
Federico~A Galatolo, Mario~GCA Cimino, and Gigliola Vaglini.
\newblock Generating images from caption and vice versa via {CLIP}-guided
  generative latent space search.
\newblock {\em arXiv preprint arXiv:2102.01645}, 2021.

\bibitem{gatys_cvpr_2016}
Leon~A Gatys, Alexander~S Ecker, and Matthias Bethge.
\newblock Image style transfer using convolutional neural networks.
\newblock In {\em Proceedings of the IEEE conference on computer vision and
  pattern recognition}, pages 2414--2423, 2016.

\bibitem{goodfellow_nips_2014}
Ian Goodfellow, Jean Pouget-Abadie, Mehdi Mirza, Bing Xu, David Warde-Farley,
  Sherjil Ozair, Aaron Courville, and Yoshua Bengio.
\newblock Generative adversarial nets.
\newblock In {\em Advances in neural information processing systems}, pages
  2672--2680, 2014.

\bibitem{heusel_nips_2017}
Martin Heusel, Hubert Ramsauer, Thomas Unterthiner, Bernhard Nessler, and Sepp
  Hochreiter.
\newblock {GANs} trained by a two time-scale update rule converge to a local
  nash equilibrium.
\newblock {\em Advances in neural information processing systems}, 30, 2017.

\bibitem{ho_nips_2020}
Jonathan Ho, Ajay Jain, and Pieter Abbeel.
\newblock Denoising diffusion probabilistic models.
\newblock {\em Advances in Neural Information Processing Systems},
  33:6840--6851, 2020.

\bibitem{ho_nipsw_2021}
Jonathan Ho and Tim Salimans.
\newblock Classifier-free diffusion guidance.
\newblock In {\em NeurIPS 2021 Workshop on Deep Generative Models and
  Downstream Applications}, 2021.

\bibitem{ho_iclrw_2022}
Jonathan Ho, Tim Salimans, Alexey~A Gritsenko, William Chan, Mohammad Norouzi,
  and David~J Fleet.
\newblock Video diffusion models.
\newblock In {\em ICLR Workshop on Deep Generative Models for Highly Structured
  Data}, 2022.

\bibitem{huang_cvpr_2017}
Gao Huang, Zhuang Liu, Laurens van~der Maaten, and Kilian~Q Weinberger.
\newblock Densely connected convolutional networks.
\newblock In {\em Proceedings of the IEEE Conference on Computer Vision and
  Pattern Recognition}, 2017.

\bibitem{huang_iccv_2017}
Xun Huang and Serge Belongie.
\newblock Arbitrary style transfer in real-time with adaptive instance
  normalization.
\newblock In {\em Proceedings of the IEEE International Conference on Computer
  Vision}, pages 1501--1510, 2017.

\bibitem{karras_nips_2021}
Tero Karras, Miika Aittala, Samuli Laine, Erik H{\"a}rk{\"o}nen, Janne
  Hellsten, Jaakko Lehtinen, and Timo Aila.
\newblock Alias-free generative adversarial networks.
\newblock {\em Advances in Neural Information Processing Systems}, 34:852--863,
  2021.

\bibitem{karras_cvpr_2019}
Tero Karras, Samuli Laine, and Timo Aila.
\newblock A style-based generator architecture for generative adversarial
  networks.
\newblock In {\em Proceedings of the IEEE/CVF conference on computer vision and
  pattern recognition}, pages 4401--4410, 2019.

\bibitem{kong_iclr_2021}
Zhifeng Kong, Wei Ping, Jiaji Huang, Kexin Zhao, and Bryan Catanzaro.
\newblock Diffwave: A versatile diffusion model for audio synthesis.
\newblock In {\em International Conference on Learning Representations}, 2021.

\bibitem{kwon_cvpr_2022}
Gihyun Kwon and Jong~Chul Ye.
\newblock {CLIPstyler}: Image style transfer with a single text condition.
\newblock In {\em Proceedings of the IEEE/CVF Conference on Computer Vision and
  Pattern Recognition}, pages 18062--18071, 2022.

\bibitem{li_cvpr_2019}
Xueting Li, Sifei Liu, Jan Kautz, and Ming-Hsuan Yang.
\newblock Learning linear transformations for fast image and video style
  transfer.
\newblock In {\em Proceedings of the IEEE/CVF Conference on Computer Vision and
  Pattern Recognition}, pages 3809--3817, 2019.

\bibitem{li_nips_2017}
Yijun Li, Chen Fang, Jimei Yang, Zhaowen Wang, Xin Lu, and Ming-Hsuan Yang.
\newblock Universal style transfer via feature transforms.
\newblock {\em Advances in neural information processing systems}, 30, 2017.

\bibitem{liao_tip_2022}
Yi-Sheng Liao and Chun-Rong Huang.
\newblock Semantic context-aware image style transfer.
\newblock {\em IEEE Transactions on Image Processing}, 31:1911--1923, 2022.

\bibitem{lin_arxiv_2021}
Junyang Lin, Rui Men, An Yang, Chang Zhou, Ming Ding, Yichang Zhang, Peng Wang,
  Ang Wang, Le Jiang, Xianyan Jia, et~al.
\newblock M6: A {Chinese} multimodal pretrainer.
\newblock {\em arXiv preprint arXiv:2103.00823}, 2021.

\bibitem{liu_iccv_2021}
Songhua Liu, Tianwei Lin, Dongliang He, Fu Li, Meiling Wang, Xin Li, Zhengxing
  Sun, Qian Li, and Errui Ding.
\newblock {AdaAttN}: Revisit attention mechanism in arbitrary neural style
  transfer.
\newblock In {\em Proceedings of the IEEE/CVF international conference on
  computer vision}, pages 6649--6658, 2021.

\bibitem{lu_arxiv_2022}
Cheng Lu, Yuhao Zhou, Fan Bao, Jianfei Chen, Chongxuan Li, and Jun Zhu.
\newblock {DPM}-solver: A fast ode solver for diffusion probabilistic model
  sampling in around 10 steps.
\newblock {\em arXiv preprint arXiv:2206.00927}, 2022.

\bibitem{lugmayr_cvpr_2022}
Andreas Lugmayr, Martin Danelljan, Andres Romero, Fisher Yu, Radu Timofte, and
  Luc Van~Gool.
\newblock {RePaint}: Inpainting using denoising diffusion probabilistic models.
\newblock In {\em Proceedings of the IEEE/CVF Conference on Computer Vision and
  Pattern Recognition}, pages 11461--11471, 2022.

\bibitem{luo_cvpr_2021}
Shitong Luo and Wei Hu.
\newblock Diffusion probabilistic models for 3d point cloud generation.
\newblock In {\em Proceedings of the IEEE/CVF Conference on Computer Vision and
  Pattern Recognition}, pages 2837--2845, 2021.

\bibitem{meng_arxiv_2021}
Chenlin Meng, Yang Song, Jiaming Song, Jiajun Wu, Jun-Yan Zhu, and Stefano
  Ermon.
\newblock Sdedit: Image synthesis and editing with stochastic differential
  equations.
\newblock {\em arXiv preprint arXiv:2108.01073}, 2021.

\bibitem{nichol_arxiv_2022}
Alex Nichol, Prafulla Dhariwal, Aditya Ramesh, Pranav Shyam, Pamela Mishkin,
  Bob McGrew, Ilya Sutskever, and Mark Chen.
\newblock {GLIDE}: Towards photorealistic image generation and editing with
  text-guided diffusion models.
\newblock {\em arXiv preprint arXiv:2112.10741}, 2022.

\bibitem{nichol_icml_2021}
Alexander~Quinn Nichol and Prafulla Dhariwal.
\newblock Improved denoising diffusion probabilistic models.
\newblock In {\em International Conference on Machine Learning}, pages
  8162--8171, 2021.

\bibitem{park_cvpr_2019}
Dae~Young Park and Kwang~Hee Lee.
\newblock Arbitrary style transfer with style-attentional networks.
\newblock In {\em Proceedings of the IEEE/CVF Conference on Computer Vision and
  Pattern Recognition}, pages 5880--5888, 2019.

\bibitem{parmar_icml_2018}
Niki Parmar, Ashish Vaswani, Jakob Uszkoreit, Lukasz Kaiser, Noam Shazeer,
  Alexander Ku, and Dustin Tran.
\newblock Image transformer.
\newblock In {\em International conference on machine learning}, pages
  4055--4064, 2018.

\bibitem{qiao_cvpr_2019}
Tingting Qiao, Jing Zhang, Duanqing Xu, and Dacheng Tao.
\newblock Mirrorgan: Learning text-to-image generation by redescription.
\newblock In {\em Proceedings of the IEEE/CVF Conference on Computer Vision and
  Pattern Recognition}, pages 1505--1514, 2019.

\bibitem{radford_icml_2021}
Alec Radford, Jong~Wook Kim, Chris Hallacy, Aditya Ramesh, Gabriel Goh,
  Sandhini Agarwal, Girish Sastry, Amanda Askell, Pamela Mishkin, Jack Clark,
  et~al.
\newblock Learning transferable visual models from natural language
  supervision.
\newblock In {\em International Conference on Machine Learning}, pages
  8748--8763, 2021.

\bibitem{ramesh_arxiv_2022}
Aditya Ramesh, Prafulla Dhariwal, Alex Nichol, Casey Chu, and Mark Chen.
\newblock Hierarchical text-conditional image generation with {CLIP} latents.
\newblock {\em arXiv preprint arXiv:2204.06125}, 2022.

\bibitem{ramesh_icml_2021}
Aditya Ramesh, Mikhail Pavlov, Gabriel Goh, Scott Gray, Chelsea Voss, Alec
  Radford, Mark Chen, and Ilya Sutskever.
\newblock Zero-shot text-to-image generation.
\newblock In {\em International Conference on Machine Learning}, pages
  8821--8831, 2021.

\bibitem{reed_icml_2016}
Scott Reed, Zeynep Akata, Xinchen Yan, Lajanugen Logeswaran, Bernt Schiele, and
  Honglak Lee.
\newblock Generative adversarial text to image synthesis.
\newblock In {\em International conference on machine learning}, pages
  1060--1069, 2016.

\bibitem{rombach_cvpr_2022}
Robin Rombach, Andreas Blattmann, Dominik Lorenz, Patrick Esser, and Bj{\"o}rn
  Ommer.
\newblock High-resolution image synthesis with latent diffusion models.
\newblock In {\em Proceedings of the IEEE/CVF Conference on Computer Vision and
  Pattern Recognition}, pages 10684--10695, 2022.

\bibitem{romero_cvpr_2022}
Andres Romero, Angela Castillo, Jose Abril-Nova, Radu Timofte, Ritwik Das,
  Sanchit Hira, Zhihong Pan, Min Zhang, Baopu Li, Dongliang He, et~al.
\newblock {NTIRE} 2022 image inpainting challenge: Report.
\newblock In {\em Proceedings of the IEEE/CVF Conference on Computer Vision and
  Pattern Recognition}, pages 1150--1182, 2022.

\bibitem{saharia_nipsw_2021}
Chitwan Saharia, William Chan, Huiwen Chang, Chris~A Lee, Jonathan Ho, Tim
  Salimans, David~J Fleet, and Mohammad Norouzi.
\newblock Palette: Image-to-image diffusion models.
\newblock In {\em NeurIPS 2021 Workshop on Deep Generative Models and
  Downstream Applications}, 2021.

\bibitem{saharia_arxiv_2022}
Chitwan Saharia, William Chan, Saurabh Saxena, Lala Li, Jay Whang, Emily
  Denton, Seyed Kamyar~Seyed Ghasemipour, Burcu~Karagol Ayan, S~Sara Mahdavi,
  Rapha~Gontijo Lopes, et~al.
\newblock Photorealistic text-to-image diffusion models with deep language
  understanding.
\newblock {\em arXiv preprint arXiv:2205.11487}, 2022.

\bibitem{saharia_arxiv_2021}
Chitwan Saharia, Jonathan Ho, William Chan, Tim Salimans, David~J Fleet, and
  Mohammad Norouzi.
\newblock Image super-resolution via iterative refinement.
\newblock {\em arXiv preprint arXiv:2104.07636}, 2021.

\bibitem{saleh_arxiv_2015}
Babak Saleh and Ahmed Elgammal.
\newblock Large-scale classification of fine-art paintings: Learning the right
  metric on the right feature.
\newblock {\em arXiv preprint arXiv:1505.00855}, 2015.

\bibitem{sohl_icml_2015}
Jascha Sohl-Dickstein, Eric Weiss, Niru Maheswaranathan, and Surya Ganguli.
\newblock Deep unsupervised learning using nonequilibrium thermodynamics.
\newblock In {\em International Conference on Machine Learning}, pages
  2256--2265, 2015.

\bibitem{song_iclr_2021}
Jiaming Song, Chenlin Meng, and Stefano Ermon.
\newblock Denoising diffusion implicit models.
\newblock In {\em International Conference on Learning Representations}, 2021.

\bibitem{song_nips_2019}
Yang Song and Stefano Ermon.
\newblock Generative modeling by estimating gradients of the data distribution.
\newblock {\em Advances in Neural Information Processing Systems}, 32, 2019.

\bibitem{song_iclr_2020}
Yang Song, Jascha Sohl-Dickstein, Diederik~P Kingma, Abhishek Kumar, Stefano
  Ermon, and Ben Poole.
\newblock Score-based generative modeling through stochastic differential
  equations.
\newblock In {\em International Conference on Learning Representations}, 2020.

\bibitem{tao_arxiv_2020}
Ming Tao, Hao Tang, Songsong Wu, Nicu Sebe, Xiao-Yuan Jing, Fei Wu, and Bingkun
  Bao.
\newblock {DF-GAN}: Deep fusion generative adversarial networks for
  text-to-image synthesis.
\newblock {\em arXiv preprint arXiv:2008.05865}, 2020.

\bibitem{tulyakov_cvpr_2018}
Sergey Tulyakov, Ming-Yu Liu, Xiaodong Yang, and Jan Kautz.
\newblock {MoCoGAN}: Decomposing motion and content for video generation.
\newblock In {\em Proceedings of the IEEE conference on computer vision and
  pattern recognition}, pages 1526--1535, 2018.

\bibitem{van_jmlr_2008}
Laurens Van~der Maaten and Geoffrey Hinton.
\newblock Visualizing data using {t-SNE}.
\newblock {\em Journal of machine learning research}, 9(11), 2008.

\bibitem{wang_arxiv_2021}
Chulin Wang, Kyongmin Yeo, Xiao Jin, Andres Codas, Levente~J Klein, and Bruce
  Elmegreen.
\newblock {S3RP}: Self-supervised super-resolution and prediction for
  advection-diffusion process.
\newblock {\em arXiv preprint arXiv:2111.04639}, 2021.

\bibitem{xu_cvpr_2018}
Tao Xu, Pengchuan Zhang, Qiuyuan Huang, Han Zhang, Zhe Gan, Xiaolei Huang, and
  Xiaodong He.
\newblock {AttnGAN}: Fine-grained text to image generation with attentional
  generative adversarial networks.
\newblock In {\em Proceedings of the IEEE conference on computer vision and
  pattern recognition}, pages 1316--1324, 2018.

\bibitem{zhang_iccv_2017}
Han Zhang, Tao Xu, Hongsheng Li, Shaoting Zhang, Xiaogang Wang, Xiaolei Huang,
  and Dimitris~N Metaxas.
\newblock {StackGAN}: Text to photo-realistic image synthesis with stacked
  generative adversarial networks.
\newblock In {\em Proceedings of the IEEE international conference on computer
  vision}, pages 5907--5915, 2017.

\end{thebibliography}

\end{document}